%% file: iclr2023_conference.tex
\documentclass{article} 
\usepackage{iclr2023_conference,times}

\input{math_commands.tex}

\usepackage{hyperref}
\usepackage{url}

\usepackage{color}
\usepackage{xcolor}
\definecolor{ugreen}{rgb}{0,0.5,0}
\definecolor{lgreen}{rgb}{0.9,1,0.8}
\definecolor{lightgray}{gray}{0.85}
\definecolor{myblack}{rgb}{0.15,0.15,0.15}
\definecolor{lyellow}{rgb}{0.54, 0.25, 0.27}
\definecolor{mypurple}{rgb}{0.6, 0.4, 0.8}

\definecolor{darkblue}{rgb}{0.0, 0.0, 0.55}
\definecolor{darkcandyapplered}{rgb}{0.64, 0.0, 0.0}
\usepackage{caption}
\usepackage{enumitem}
\usepackage{amsmath, amssymb}
\usepackage{amsfonts} 
\usepackage{pifont} 
\usepackage{pgfplots}
\usepackage{tikz}
\usepackage{subfig}
\usetikzlibrary{backgrounds,fit} 
\usetikzlibrary{shapes,arrows,shadows}
\usetikzlibrary{patterns}
\usetikzlibrary{shapes.geometric} 
\usetikzlibrary{decorations.pathreplacing} 
\usetikzlibrary{calc}
\usepackage{array,multirow}
\usepackage{array} 
\usepackage{booktabs} 
\usepackage{arydshln} 
\usepackage{diagbox} 
\newcommand{\PreserveBackslash}[1]{\let\temp=\\#1\let\\=\temp} 
\newcolumntype{C}[1]{>{\PreserveBackslash\centering}p{#1}}
\newcolumntype{R}[1]{>{\PreserveBackslash\raggedleft}p{#1}}
\newcolumntype{L}[1]{>{\PreserveBackslash\raggedright}p{#1}}
\newcolumntype{M}[1]{ >{\centering\arraybackslash}m{#1}}
\usepackage{cleveref}
\usepackage{xspace}
\usepackage{algorithm,algorithmicx,algpseudocode}
\usepackage{multicol}
\usepackage{bm}
\usepackage{enumitem}

\usepackage{wrapfig}

\pgfplotsset{compat=1.16}

\usepackage{verbatim}

\usepackage{ulem} 

\newlength{\vseg}
\newlength{\hseg}
\newlength{\wnode}
\newlength{\hnode}
\newcommand{\bos}{[\texttt{BOS}]\xspace}
\newcommand{\shortbos}{[\texttt{B}]\xspace}

\newcommand{\eos}{[\texttt{EOS}]\xspace}
\newcommand{\shorteos}{[\texttt{E}]\xspace}
\newcommand{\pad}{[\texttt{PAD}]\xspace}
\newcommand{\shortpad}{[\texttt{P}]\xspace}
\newcommand{\mask}{[\texttt{MASK}]\xspace}
\newcommand{\shortmask}{[\texttt{M}]\xspace}

\newcommand{\spe}[1]{[\texttt{#1}]\xspace}

\newcommand{\entode}{En$\rightarrow$De\xspace}

\newcommand{\entoro}{En$\rightarrow$Ro\xspace}
\newcommand{\rotoen}{Ro$\rightarrow$En\xspace}
\newcommand{\enbothde}{En$\leftrightarrow$De\xspace}
\newcommand{\enbothro}{En$\leftrightarrow$Ro\xspace}
\newcommand{\zhtoen}{Zh$\rightarrow$En\xspace}

\title{Hybrid-Regressive Neural Machine Translation}

\author{Antiquus S.~Hippocampus, Natalia Cerebro \& Amelie P. Amygdale \thanks{ Use footnote for providing further information
about author (webpage, alternative address)---\emph{not} for acknowledging
funding agencies.  Funding acknowledgements go at the end of the paper.} \\
Department of Computer Science\\
Cranberry-Lemon University\\
Pittsburgh, PA 15213, USA \\
\texttt{\{hippo,brain,jen\}@cs.cranberry-lemon.edu} \\
\And
Ji Q. Ren \& Yevgeny LeNet \\
Department of Computational Neuroscience \\
University of the Witwatersrand \\
Joburg, South Africa \\
\texttt{\{robot,net\}@wits.ac.za} \\
\AND
Coauthor \\
Affiliation \\
Address \\
\texttt{email}
}

\begin{document}

\maketitle

\begin{abstract}

In this work, we empirically confirm that non-autoregressive translation with an iterative refinement mechanism (IR-NAT) suffers from poor acceleration robustness because it is more sensitive to decoding batch size and computing device setting than autoregressive translation (AT). 
Inspired by it, we attempt to investigate how to combine the strengths of autoregressive and non-autoregressive translation paradigms better. To this end, we demonstrate through synthetic experiments that prompting a small number of AT's predictions can promote one-shot non-autoregressive translation to achieve the equivalent performance of IR-NAT. Following this line, we propose a new two-stage translation prototype called hybrid-regressive translation (HRT). Specifically, HRT first generates discontinuous sequences via autoregression (e.g., make a prediction every $k$ tokens, $k>1$) and then fills in all previously skipped tokens at once in a non-autoregressive manner. We also propose a bag of techniques to effectively and efficiently train HRT without adding any model parameters. HRT achieves the state-of-the-art BLEU score of 28.49 on the WMT \entode task and is at least 1.5x faster than AT, regardless of batch size and device. In addition, another bonus of HRT is that it successfully inherits the good characteristics of AT in the deep-encoder-shallow-decoder architecture. Concretely, compared to the vanilla HRT with a 6-layer encoder and 6-layer decoder, the inference speed of HRT with a 12-layer encoder and 1-layer decoder is further doubled on both GPU and CPU without BLEU loss.
\footnote{We will release the source code once accepted.}

\end{abstract}

\section{Introduction}

Autoregressive translation (AT) such as Transformer has been the \textit{de facto} standard for Neural Machine Translation (NMT) \cite{vaswani2017attention}. However, AT predicts only \textit{one} target word each time, resulting in a slow inference speed. To address this problem, non-autoregressive translation (NAT) attempts to generate the entire target sequence in parallel in one shot, assuming that the generation of target tokens is conditional independent \cite{gu2018nonautoregressive}. While efficient, one-shot NAT suffers from severe translation quality degradation.
How to achieve a better balance between inference speed and translation quality is still an active field for NAT \cite{wang2018semi,ran-etal-2020-learning,qian-etal-2021-glancing,huang2022improving}.

The iterative refinement mechanism first introduced by \citet{lee2018deterministic}, is one of the most successful approaches to this issue and has been adopted by several leading systems \cite{ghazvininejad-etal-2019-mask,kasai2020non,guo2020jointly,saharia2020non,geng-etal-2021-learning,huang2022improving}.  
Specifically, iterative refinement-based NAT (abbreviated as IR-NAT), also known as multi-shot NAT, ``thinks more" than one-shot NAT: IR-NAT takes the translation hypothesis from the previous iteration as a reference and regularly polishes the new translation. The iteration stops when reaching the predefined iteration count $I$ or no translation changed. A larger $I$ generally improves translation accuracy while facing the risk of speedup degradation \cite{kasai2020deep}.

In this work, we devote ourselves to understanding the limitations of IR-NAT and attempting to build a new fast-and-accurate translation paradigm beyond it. 
Our contributions are the following:
\begin{itemize}
	\item
	We comprehensively study the acceleration robustness problem in IR-NAT and extend the finding of \cite{kaiser2018fast} that IR-NAT is more sensitive not only to the decoding batch size but also to the computing device compared with AT.
	For example, when the decoding batch size is 1/8/16/32, the ten-iteration non-autoregressive model achieves 1.7x/1.2x/0.7x/0.4x inference speed of the AT model on GPU, respectively. However, when switching to CPU, the relative speed ratio drops to 0.8x/0.4x/0.3x/0.3x. These results prove that the two translation paradigms are well complementary to each other.
	
	\item 
	We design synthetic experiments to investigate how much target context (i.e., the number of target tokens) is sufficient for one-shot NAT to rival multi-shot NAT. 
	The value of the answer is that we could build the desired target context more cheaply, replacing expensive multiple iterations.
	Specifically, given a well-trained CMLM model, we notice that under the appropriate masking strategy, even if 70\% of AT translations are masked, the remaining target context can help the CMLM$_1$ with greedy search compete with the standard CMLM$_{10}$ with beam search (see Figure~\ref{fig:syn_exp}). To our best knowledge, we are the first to study the masking rate issue in the inference phase of NAT.
	
	\item 
	Inspired by the observations above, we proposed a novel two-stage translation prototype -- hybrid-regressive translation (HRT), to incorporate the advantages of AT and NAT. Concretely, HRT first uses an autoregressive decoder to generate a discontinuous target sequence with the interval $k$ ($k>1$). Then, HRT fills the remaining slots at once in a lightweight non-autoregressive manner. We propose to use a multi-task learning framework enhanced by curriculum learning and mixed distillation for effective and efficient training without adding any model parameters. 
    
    \item
    Experimental results on WMT \enbothro, \enbothde, and NIST \zhtoen show that HRT significantly outperforms prior work combining AT and NAT and has competitive BLEU with state-of-the-art IR-NAT models. Specifically, HRT achieves a BLEU score of 28.49 on the WMT \entode task and is robust 1.5x faster than AT regardless of batch size and device. Moreover, HRT equipped with deep-encoder-shallow-decoder architecture achieves up to 4x/3x acceleration on GPU/CPU, respectively, without BLEU loss.
\end{itemize}

\section{Background}
Given a source sentence $\vx=\{x_1, x_2, \ldots, x_M\}$ and a target sentence $\vy=\{y_1, y_2, \ldots, y_N\}$, there are several ways to model $P(\vy|\vx)$:

\paragraph{Autoregressive Translation (AT) }
AT is the dominant approach in NMT, which decomposes $P(\vy|\vx)$ by chain rules:
$P(\vy|\vx) = \prod_{t=1}^{N} P(y_t|\vx,y_{<t})$,
where $y_{<t}$ denotes the generated prefix translation before time step $t$. 
However, autoregressive models have to wait for the generation of $y_{t-1}$ before predicting $y_{t}$, which hinders the parallelism over the target sequence.

\paragraph{Non-Autoregressive Translation (NAT) } NAT allows generating all target tokens simultaneously \cite{gu2018nonautoregressive}. NAT replaces $y_{<t}$ with target-independent input $\vz$ and rewrites as:
$P(\vy|\vx) = P(N|\vx) \times \prod_{t=1}^{N} P(y_t|\vx,\vz)$.
We can model $\vz$ as source embedding \citep{gu2018nonautoregressive,guo2019non}, reordered source sentence \citep{ran2019guiding}, latent variable \citep{ma2019flowseq,shu2019latent} etc.

\paragraph{Iterative Refinement based Non-Autoregressive Translation (IR-NAT) }
IR-NAT extends the traditional one-shot NAT by introducing an iterative refinement mechanism \citep{lee2018deterministic}. We choose CMLM as the IR-NAT representative in this work due to its excellent performance and simplification \cite{ghazvininejad-etal-2019-mask}. During training, CMLM randomly masks a fraction of tokens on $\vy$ as the alternative to $\vz$ and is trained as a conditional masked language model \cite{devlin-etal-2019-bert}. Denote $\vy^m$/$\vy^r$ as the masked/residual tokens of $\vy$, then we have:
$P(\vy|\vx) = \prod_{t=1}^{|\vy^m|} P(\vy_t^m|\vx, \vy^r)$.
At inference, CMLM deterministically masks tokens from the hypothesis in the previous iteration $\hat{\vy}^{(i-1)}$ according to prediction confidences.
This process is repeated until $\hat{\vy}^{(i-1)}$=$\hat{\vy}^{(i)}$ or $i$ reaches the maximum iteration count.

\section{Acceleration Robustness Problem}
\label{sec:problem}

\begin{table}[t]
\begin{tabular}{M{.63\textwidth}M{.32\textwidth}}
     
     \begin{minipage}[t]{.63\textwidth}
     
	\centering
	\resizebox{.95\textwidth}{!}
	{
		\begin{tikzpicture}
			\begin{axis}[domain  = 0.97:2.3,
				width=\textwidth,
				height=.5\textwidth,
				minor x tick num=4,
				minor y tick num=4,
				samples = 100,
				xmin    = -1,
				xmax    = 35,
				ymin    = -4,
				ymax    = 10,
				ytick   = \empty,
				xlabel  = {\scriptsize{Decoding Batch Size}},
				ylabel  = {\scriptsize{Relative Speedup Ratio ($\alpha$)}},
				ytick={-3, 1, 5, 10},
				yticklabels={0, 1, 5, 10},
				tick label style={font=\scriptsize},
				xlabel near ticks,
				ylabel near ticks,
				grid=major,
				legend style={at={(0.65,+.6)},
					anchor=south,legend columns=2},
				]
				
				\draw[dotted, thick] (axis cs: -10, 1 )-- (axis cs: 69, 1);
				
				\addplot [darkblue, mark=diamond*, line width=1pt, smooth] coordinates {
					(1, 9.58836689) (8, 4.784722222) (16, 2.779467681) (32, 1.530534351)
				};
				\addlegendentry{\scriptsize{CMLM$_{1}$}};
				
				\addplot [ugreen, mark=otimes*, line width=1pt, smooth] coordinates {
				    (1, 3.835346756) (8, 2.327702703) (16, 1.405769231) (32, 0.765267176*4-3)
				};
				\addlegendentry{\scriptsize{CMLM$_{4}$}};
				
				\addplot [darkcandyapplered, mark=star, line width=1pt, smooth] coordinates {
				    (1, 1.739801096) (8, 1.187931034) (16, 0.688972667*4-3) (32, 0.381904762*4-3)
				};
	        	\addlegendentry{\scriptsize{CMLM$_{10}$}};
				
				\addplot [darkblue, dashed, mark=diamond*, line width=1pt, smooth] coordinates {
					(1, 5.393150024) (8, 2.707758621) (16, 2.523565574) (32, 2.343422584)
				};
				
				\addplot [ugreen, dashed, mark=otimes*, line width=1pt, smooth] coordinates {
					(1, 1.7603527) (8, 0.919227392*4-3) (16, 0.746816252*4-3) (32, 0.630244208*4-3)
				};
				
				\addplot [darkcandyapplered, dashed, mark=star, line width=1pt, smooth] coordinates {
				    (1, 0.803796103*4-3) (8, 0.383002073*4-3) (16, 0.312048651*4-3) (32, 0.259106706*4-3)
				};
				
			\end{axis}
		\end{tikzpicture}
	}
	\vspace{-.5em}
     \end{minipage}
     &
    \begin{minipage}[t]{.32\textwidth}
    
	
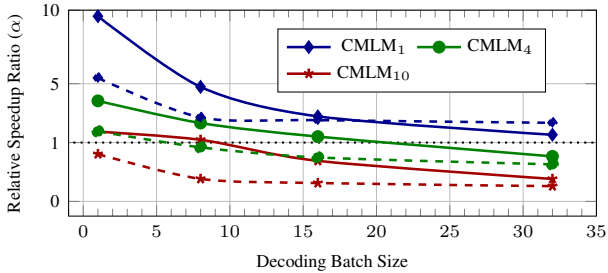
\captionof{figure}{Relative speedup ratio ($\alpha$) compared CMLM with AT on GPU (solid) and CPU (dashed). We test different batch sizes from $\{1, 8, 16, 32\}$. 
	$\alpha < 1$ denotes that CMLM runs slower than AT. 
	CMLM$_1$ can be approximately regarded as the representative of one-shot NAT. }
	\label{fig:batch_wrt_speed}
	\vspace{-.5em}
    \end{minipage}
    
     \\
     
\end{tabular}
\end{table}

In this section, we attempt to comprehensively understand the inference acceleration robustness problem in IR-NAT. Without loss of generality, we take CMLM as the agency of IR-NAT.\footnote{From the perspective of inference speed, we note that most one-shot NAT models are closed to CMLM$_1$. Especially, existing one-shot NAT models with CTC loss, such as GLAT, and Fully-NAT, are theoretically slower than CMLM$_1$ because they require a long enough target sequence for inference.}

\paragraph{Problem Description }
The inference overhead of the autoregressive translation model mainly concentrates on the decoder side\cite{hu-etal-2020-niutrans}. Suppose that the decoder's computational cost is proportional to the size of its input tensor $(B, N, H)$, where $B$ is the batch size, $N$ is the target sequence length, and $H$ is the network dimension. For convenience, we omit $H$ due to its invariance in NAT and AT. Thus, the total cost of AT model is about $C_{at} \propto N \times \mathcal{O}(B \times 1)$ \footnote{Though the decoder self-attention module considers the previous $i$ tokens, we omit it here for the sake of clarity.}. Likely, the cost of $I$-iteration NAT is $C_{nat} \propto I \times \mathcal{O}(B \times N)$. Given a fixed test set, We use $\mathcal{T}_{D}(\cdot)$ to represent the translation time 
on computing device $D$. In this way, we get the relative speedup ratio $\alpha$ between $I$-iteration NAT and AT by:
\begin{equation}
	\label{eq:alpha}
\alpha = \frac{\mathcal{T}_{D}(C_{at})}{\mathcal{T}_{D}(C_{nat)}} \propto \frac{N}{I} \times \mathcal{E}(B, D),
\end{equation} 
where $\mathcal{E}(B, D)$=$\frac{\mathcal{T}_{D}(\mathcal{O}(B \times 1))}{\mathcal{T}_{D}(\mathcal{O}(B \times N))} \leq1$, denotes the parallel computation efficiency over sequence under batch size $B$ and device $D$.
When fixing $N$ and $I$, $\alpha$ is completely determined by $\mathcal{E}(B, D)$. 
We note that most previous NAT studies only report the inference speed with $D$=GPU and $B$=1, without considering cases when $B$ or $D$ change.

\paragraph{Setup}
To systematically investigate this problem, we compared the actual inference speed of CMLM \footnote{We use the officially released CMLM models: \url{https://github.com/facebookresearch/Mask-Predict}.} and AT under varying environments, including batch size $B \in \{1, 8, 16, 32\}$, device $D \in \{\textrm{GPU}, \textrm{CPU}\}$  \footnote{
Unless otherwise stated, we use 2080Ti GPU and Intel Xeon(R) E5-2683 v4 CPU in this work. 
}, and the number of iteration $I \in \{1, 4, 10\}$. 
We use a beam size of 5 in all experiments.
We test inference speed on the widely used WMT \entode \textit{newstest2014} test set and report the average results over five runs (see Appendix~\ref{appendix:variant_speedup} for details).

\paragraph{Results }
We plot the curve of relative speedup ratio ($\alpha$) in Figure~\ref{fig:batch_wrt_speed}, where we can see that: 
\begin{enumerate}[label=\roman*.]
	\item
    $\alpha$ decreases as decoding batch size increases regardless of the number of iterations, which is in line with \cite{kasai2020deep}.
    \item
    $\alpha$ on CPU generally performs worse than that on GPU, except using one iteration.
    \item
    The benefit of non-autoregressive decoding is more prone to disappear for larger $I$.
\end{enumerate}
More concretely, the ten-iteration non-autoregressive model is 70\% faster than the autoregressive counterpart when decoding a single sentence on GPU. In contrast, the IR-NAT model only reaches 30\% inference speed of the AT model when switching to batches of 32 on CPU.
These results indicate that the two translation paradigms enjoy different decoding setups and complement each other. Therefore, combining the advantages of AT and NAT could be an effective way for robust acceleration.

\section{Synthetic Experiments}
\label{sec:syn_exp}

According to Equation~\ref{eq:alpha}, we know that reducing the iteration count $I$ helps to increase $\alpha$. Recalling the refinement process of IR-NAT, we think the essence of multiple iterations is to provide the decoder with a good enough target context (deterministic target tokens).
Thus, an unexplored question raises that \textit{how many target tokens need to be provided to make one-shot NAT compete with IR-NAT}? In this section, we try to answer this question through synthetic experiments on WMT \entoro and \entode. Specifically, we control the size of the target context by masking the partial translations generated by a pre-trained AT model. Then we use a pre-trained CMLM model to predict these masks. Finally, we observe the BLEU score curves under different masking rates.

\begin{figure*}[t]
	\begin{center}
		\renewcommand\arraystretch{0.0}
		\setlength{\tabcolsep}{1pt}
		\begin{tabular}{c c}
			
			\subfloat
			{
				\begin{tikzpicture}{baseline}
					\begin{axis}[
						ylabel near ticks,
						minor x tick num=1,
						minor y tick num=4,
						width=.45\textwidth,
						height=.23\textwidth,
						legend style={at={(0.5,+1.05)},
							anchor=south,legend columns=-1},
						xlabel={\scriptsize{masking rate p$_{mask}$}},
						ylabel={\scriptsize{BLEU Score (\%)}},
						ylabel style={yshift=-0em},xlabel style={yshift=0.0em},
						tick label style={font=\scriptsize},
						yticklabel style={/pgf/number format/precision=0,/pgf/number format/fixed zerofill},
						xmin=0,xmax=1.0,xtick={.0,.2,.4,.6,.8,1.0},
						xmajorgrids,
						]
						
						\addplot+[mark=o, mark options={scale=1}, line width=0.5pt] coordinates {(.1, 34.16) (.2, 33.82) (.3, 33.31) (.4, 32.39) (.5, 31.32) (.6, 30.26) (.7, 28.95) (.8, 27.22) (.9, 25.63) (1., 25.07)};
						\addlegendentry{\scriptsize{head}};
						
						\addplot+[mark=square, mark options={scale=1},line width=0.5pt] coordinates {(.1, 34.24) (.2, 34.05) (.3, 33.6) (.4, 32.65) (.5, 31.52) (.6, 30.49) (.7, 29.47) (.8, 28.02) (.9, 26.33) (1., 25.07)};
						
						\addlegendentry{\scriptsize{tail}};
						
						\addplot+[mark options={scale=1},mark=diamond, line width=0.5pt] coordinates {(.1, 34.23) (.2, 34.15) (.3, 34.09) (.4, 33.89) (.5, 33.73) (.6, 33.47) (.7, 33.11) (.8, 32.48) (.9, 30.96) (1., 25.07)};
						\addlegendentry{\scriptsize{random}};
						
						\addplot+[mark options={scale=1},line width=0.5pt] coordinates {(.5, 33.88) (.66, 33.56) (.75, 33.25)};
						\addlegendentry{\scriptsize{chunk}};
						
						
						\addplot[dashed,domain=0:1., lyellow, line width=0.5pt]
						{25.36};
						\node at (axis cs:0.1,25.0) [anchor=south west, color=red] {\scriptsize{b=1, I=1}};
						
						\addplot[dashed,domain=0:1., lyellow, line width=0.5pt]{27.32};
						\node at (axis cs:0.1,27.0) [anchor=south west, color=red] {\scriptsize{b=5, I=1}};
						
						\addplot[dashed,domain=0:1., lyellow, line width=0.5pt]{32.75};
						\node at (axis cs:0.1,32.7) [anchor=north west, color=red] {\scriptsize{b=1, I=10}};
						
						\addplot[dashed,domain=0:1., lyellow, line width=0.5pt]{33.08};
						\node at (axis cs:0.85,32.85) [anchor=south, color=red] {\scriptsize{b=5, I=10}};
						
					\end{axis}
				\end{tikzpicture}
			}
			\label{fig:syn_exp_en2ro}
			
			&
			\subfloat 
			{
				\begin{tikzpicture}{baseline}
					\begin{axis}[
						ylabel near ticks,
						minor x tick num=1,
						minor y tick num=4,
						width=.45\textwidth,
						height=.23\textwidth,
						legend style={at={(0.5,+1.05)},
							anchor=south,legend columns=-1},
						xlabel={\scriptsize{masking rate p$_{mask}$}},
						ylabel={\scriptsize{BLEU Score (\%)}},
						tick label style={font=\scriptsize},
						ylabel style={yshift=-0em},xlabel style={yshift=0.0em},
						yticklabel style={/pgf/number format/precision=0,/pgf/number format/fixed zerofill},
						ymin=15, ymax=30, 
						xmin=0,xmax=1.0,xtick={.0,.2,.4,.6,.8,1.0},
						xmajorgrids,
						]
						
						\addplot+[mark=o, line width=0.5pt] coordinates {(.1, 27.32) (.2, 27.16) (.3, 26.65) (.4, 25.86) (.5, 24.78) (.6, 23.52) (.7, 22.14) (.8, 20.55) (.9, 18.66) (1., 17.11)};
						\addlegendentry{\scriptsize{head}};
						
						\addplot+[mark=square, line width=0.5pt] coordinates 
						{(.1, 27.32) (.2, 27.16) (.3, 26.63) (.4, 25.96) (.5, 24.94) (.6, 23.76) (.7, 22.28) (.8, 20.61) (.9, 18.89) (1., 17.11)};
						\addlegendentry{\scriptsize{tail}};
						
						\addplot+[mark=diamond,line width=0.5pt] coordinates {(.1, 27.30) (.2, 27.28) (.3, 27.25) (.4, 27.18) (.5, 27.11) (.6, 26.99) (.7, 26.68) (.8, 26.08) (.9, 24.63) (1., 17.11)};
						\addlegendentry{\scriptsize{random}};
						
						\addplot+[line width=0.5pt] coordinates {(.5, 27.31) (.66, 27.1) (.75, 27.03)};
						\addlegendentry{\scriptsize{chunk}};
						
						\addplot[dashed,domain=0:1., lyellow, line width=0.5pt]
						{16.84};
						\node at (axis cs:0.1,15.3) [anchor=south west, color=red] {\scriptsize{b=1, I=1}};
						
						\addplot[dashed,domain=0:1., lyellow, line width=0.5pt]{18.06};
						\node at (axis cs:0.1,17.6) [anchor=south west, color=red] {\scriptsize{b=5, I=1}};
						
						\addplot[dashed,domain=0:1., lyellow, line width=0.5pt]{26.56};
						\node at (axis cs:0.1,26.5) [anchor=north west, color=red] {\scriptsize{b=1, I=10}};
						
						\addplot[dashed,domain=0:1., lyellow, line width=0.5pt]{27.03};
						\node at (axis cs:0.9,26.5) [anchor=south, color=red] {\scriptsize{b=5, I=10}};
					\end{axis}	
				\end{tikzpicture}
			}
			\label{fig:syn_exp_en2de}
		\end{tabular}
	\end{center}
	
	\begin{center}
		\vspace{-1em}
		\caption{ Comparison of four masking strategies \{\texttt{HEAD}, \texttt{TAIL}, \texttt{RANDOM}, \texttt{CHUNK}\} in synthetic experiments on WMT \entoro (Left) and \entode (Right) test sets.  For \texttt{CHUNK}, we test the chunk size from \{2, 3, 4\}. 
			Dashed lines are the official CMLM scores. $b$ stands for ``beam size,'' and $I$ stands for ``the number of iterations''.}
		\label{fig:syn_exp}
		\vspace{-1.0em}
	\end{center}
\end{figure*}
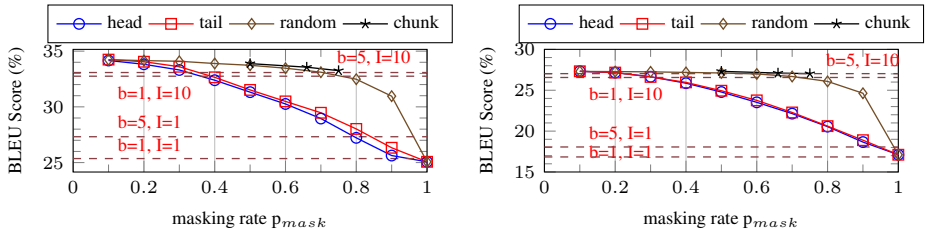

\paragraph{Models}	
We use the official CMLM models. Since the authors did not release the AT baselines, we used the same data to retrain AT models with the standard Transformer-Base configuration \citep{vaswani2017attention} and obtain comparable performance with theirs (see Appendix~\ref{appendix:syn_exp} for details).

\paragraph{Decoding} 
AT models decode with beam sizes of 5 on both tasks.
Then we replace a certain percentage of AT tokens with \mask and feed them to CMLM.
The used CMLM model only iterates once with beam size 1. We substitute all [\texttt{MASK}]s with CMLM's predictions to obtain the final translation. We report case-sensitive tokenized BLEU scores by \textit{multi-bleu.perl}.

\paragraph{Mask Strategies}

We tested four strategies to mask AT results: \texttt{HEAD}, \texttt{TAIL}, \texttt{RANDOM} and \texttt{CHUNK}. Given the masking rate p$_{mask}$ and the translation length $N$, the number of masked tokens is N$_{mask}$=max(1, $\lfloor N \times $p$_{mask} \rfloor$). Then \texttt{HEAD}/\texttt{TAIL} always masks the first/last N$_{mask}$ tokens, while \texttt{RANDOM} masks the translation randomly. \texttt{CHUNK} is slightly different from the above strategies. It first divides the target sentence into $C$ chunks, where $C=\textrm{Ceil}(N/k)$ and $k$ is the chunk size. Then in each chunk, we retain the first token but mask other $k-1$ tokens. Thus, the actual masking rate in \texttt{CHUNK} is $1-1/k$ instead of p$_{mask}$. We ran \texttt{RANDOM} three times with different seeds to exclude randomness and report the average results.

\paragraph{Results} 
The experimental results are illustrated in Figure~\ref{fig:syn_exp}, where we can see that \texttt{CHUNK} is moderately but consistently superior to \texttt{RANDOM} and both of them significantly outperform \texttt{HEAD} and \texttt{TAIL}. We attribute the success of \texttt{CHUNK} to two aspects: the use of
(1) bidirectional context \citep{devlin-etal-2019-bert} (vs. \texttt{HEAD} and \texttt{TAIL}); 
(2) uniformly distributed deterministic tokens (vs. \texttt{RANDOM}) \footnote{\texttt{CHUNK} can guarantee that each masked token (except the last $k$-1 ones in the sequence) can meet two deterministic tokens within the window size of $k$. However, in extreme cases, \texttt{RANDOM} may degrade into \texttt{HEAD}/\texttt{TAIL}.}.
In addition, when using the \texttt{CHUNK} strategy, we find that exposing 30\% AT tokens as the input of the decoder is sufficient to make our CMLM$_1$(beam=1) compete with the official CMLM$_{10}$(beam=5), which indicates the importance of a good partial target context.

\section{Hybrid-Regressive Translation}
\label{sec:hrt}

\noindent  
Inspired by the observations above, we propose a novel two-stage translation paradigm called hybrid-regressive translation (HRT) to imitate the process of \texttt{CHUNK}. Briefly speaking, HRT autoregressively generates a discontinuous sequence with chunk size $k$ (stage \uppercase\expandafter{\romannumeral1}), and then non-autoregressively fills the skipped tokens (stage \uppercase\expandafter{\romannumeral2}). 

\subsection{Architecture}
\label{sec:hrt_model}

\paragraph{Overview} 
HRT consists of three components: encoder, Skip-AT decoder (for stage \uppercase\expandafter{\romannumeral1}), and Skip-CMLM decoder (for stage \uppercase\expandafter{\romannumeral2}). All components adopt the Transformer architecture \citep{vaswani2017attention}. 
The two decoders have the same network structure, and we share them to make the parameter size of HRT the same as the vanilla Transformer.
The only difference between the two decoders lies in the masking pattern in self-attention: The Skip-AT decoder masks future tokens to guarantee strict left-to-right generation like autoregressive Transformer. In contrast, the Skip-CMLM decoder eliminates it to leverage the bi-directional context like CMLM \cite{ghazvininejad-etal-2019-mask}. 

\paragraph{No Target Length Predictor} 
Thanks to Skip-AT, we can obtain the translation length as its by-product: $N_{nat}$=$k\times N_{at}$, where $N_{at}$ is the sequence length produced by Skip-AT 
. Compared to most NAT models that jointly train the translation length predictor and the translation model, our approach benefits in two aspects: (1) There is no need to carefully tune the weighting coefficient between the sentence-level length prediction loss and the word-level target token prediction loss; (2) The length predicted by Skip-AT could be more accurate because it can access the already generated sequence information.

\subsection{Training}
\label{sec:hrt_train}

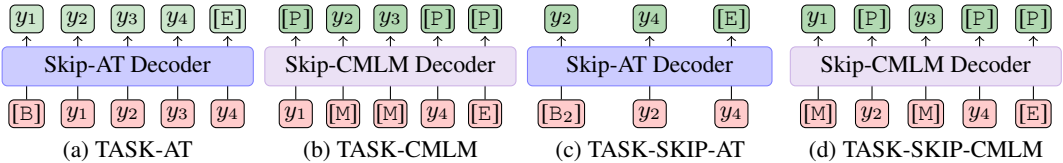
\begin{figure*}[t]
	\begin{center}
		\setlength{\tabcolsep}{2pt}

		\begin{tabular}{C{.24\textwidth}C{.24\textwidth}C{.24\textwidth}C{.24\textwidth}}
			
		\subfloat [\small{TASK-AT}] 
		{
		\begin{tikzpicture}
		\begin{scope}
		\setlength{\vseg}{2.5em}
		\setlength{\hseg}{0.6em}
		\setlength{\wnode}{2.5em}
		\setlength{\hnode}{1.5em}
		\tikzstyle{inputnode} = [rectangle, draw, thin, rounded corners=2pt, inner sep=1pt, fill=red!20, minimum height=0.7\hnode, minimum width=0.5\wnode]
		\tikzstyle{decnode} = [fill=blue!20,minimum width=3.8*\wnode, draw=blue!50,rounded corners=2pt]
		\tikzstyle{outputnode} = [rectangle, draw, thin, rounded corners=2pt, inner sep=1pt, fill=ugreen!20, minimum height=0.7\hnode, minimum width=0.5\wnode]
		
		\node [inputnode, anchor=west] (i1) at (0,0) {\footnotesize{\shortbos} };
		\node [outputnode, anchor=south] (o1) at ([yshift=\vseg] i1.north) {\footnotesize{$y_1$} };
		\draw [->,thin] (i1.north) -- ([yshift=-1pt] o1.south);
		
		\foreach \p/\i/\x/\y in {1/2/$y_1$/$y_2$,2/3/$y_2$/$y_3$,3/4/$y_3$/$y_4$,4/5/$y_4$/\shorteos}
		{
			\node [inputnode, anchor=west] (i\i) at ([xshift=\hseg] i\p.east) {\footnotesize{\x} };	
			\node [outputnode, anchor=south] (o\i) at ([yshift=\vseg] i\i.north) {\footnotesize{\y} };
		    \draw [->,thin] (i\i.north) -- ([yshift=-1pt] o\i.south);
		}
		\node[decnode] at ([yshift=0.5\vseg] i3.north) {\footnotesize{Skip-AT Decoder}};
		\end{scope}			
		\end{tikzpicture}	
		} &
	    \subfloat [\small{TASK-CMLM}]
	    {
		\begin{tikzpicture}
		\begin{scope}
		\setlength{\vseg}{2.5em}
		\setlength{\hseg}{.5em}
		\setlength{\wnode}{2.5em}
		\setlength{\hnode}{1.5em}
		\tikzstyle{inputnode} = [rectangle, draw, thin, rounded corners=2pt, inner sep=0pt, fill=red!20, minimum height=0.7\hnode, minimum width=0.5\wnode]
		\tikzstyle{decnode} = [fill=mypurple!20,minimum width=3.8*\wnode, draw=mypurple!50,rounded corners=2pt]
		\tikzstyle{outputnode} = [rectangle, draw, thin, rounded corners=2pt, inner sep=0pt, fill=ugreen!30, minimum height=0.7\hnode, minimum width=0.5\wnode]
		
		\node [inputnode, anchor=west] (i1) at (0,0) {\footnotesize{$y_1$} };
		\node [outputnode, anchor=south] (o1) at ([yshift=\vseg] i1.north) {\footnotesize{\spe{P}} };
		\draw [->,thin] (i1.north) -- ([yshift=-1pt] o1.south);
		
		\foreach \p/\i/\x/\y in {1/2/\shortmask/$y_2$,2/3/\shortmask/$y_3$,3/4/$y_4$/\shortpad,4/5/\shorteos/\shortpad}
		{
			\node [inputnode, anchor=west] (i\i) at ([xshift=\hseg] i\p.east) {\footnotesize{\x} };					
			\node [outputnode, anchor=south] (o\i) at ([yshift=\vseg] i\i.north) {\footnotesize{\y} };
		
		    \draw [->,thin] (i\i.north) -- ([yshift=-1pt] o\i.south);
		}
		\node[decnode] at ([yshift=0.5\vseg] i3.north) {\footnotesize{Skip-CMLM Decoder}};
		
		\end{scope}			
		\end{tikzpicture}

		} &
		\subfloat [\small{TASK-SKIP-AT}]
		{
		\begin{tikzpicture}
		\begin{scope}
		\setlength{\vseg}{2.5em}
		\setlength{\hseg}{.9em}
		\setlength{\wnode}{2.5em}
		\setlength{\hnode}{1.5em}
		\tikzstyle{inputnode} = [rectangle, draw, thin, rounded corners=2pt, inner sep=1pt, fill=red!20, minimum height=0.7\hnode, minimum width=0.5\wnode]
		\tikzstyle{decnode} = [fill=blue!20,minimum width=3.7*\wnode, draw=blue!50,rounded corners=2pt]
		\tikzstyle{outputnode} = [rectangle, draw, thin, rounded corners=2pt, inner sep=1pt, fill=ugreen!30, minimum height=0.7\hnode, minimum width=0.5\wnode]
		
		\node [inputnode, anchor=west] (i1) at (0,0) {\footnotesize{\spe{B$_2$}} };
		\node [outputnode, anchor=south] (o1) at ([yshift=\vseg] i1.north) {\footnotesize{$y_2$} };
		\draw [->,thin] (i1.north) -- ([yshift=-1pt] o1.south);
		
		\foreach \p/\i/\x/\y in {1/2/$y_2$/$y_4$,2/3/$y_4$/\shorteos}
		{
			\node [inputnode, anchor=west] (i\i) at ([xshift=2\hseg] i\p.east) {\footnotesize{\x} };					
			\node [outputnode, anchor=south] (o\i) at ([yshift=\vseg] i\i.north) {\footnotesize{\y} };
		
		    \draw [->,thin] (i\i.north) -- ([yshift=-1pt] o\i.south);
		}
		\node[decnode] at ([yshift=0.5\vseg] i2.north) {\footnotesize{Skip-AT Decoder}};
		
		\end{scope}			
		\end{tikzpicture}

		} &
			
		\subfloat [\small{TASK-SKIP-CMLM}]
		{
		\begin{tikzpicture}
		\begin{scope}
		\setlength{\vseg}{2.5em}
		\setlength{\hseg}{.7em}
		\setlength{\wnode}{2.5em}
		\setlength{\hnode}{1.5em}
		\tikzstyle{inputnode} = [rectangle, draw, thin, rounded corners=2pt, inner sep=1pt, fill=red!20, minimum height=0.7\hnode, minimum width=0.5\wnode]
		\tikzstyle{decnode} = [fill=mypurple!20,minimum width=4.1*\wnode, draw=mypurple!50,rounded corners=2pt]
		\tikzstyle{outputnode} = [rectangle, draw, thin, rounded corners=2pt, inner sep=1pt, fill=ugreen!30, minimum height=0.7\hnode, minimum width=0.5\wnode]
		
		\node [inputnode, anchor=west] (i1) at (0,0) {\footnotesize{\spe{M}} };
		\node [outputnode, anchor=south] (o1) at ([yshift=\vseg] i1.north) {\footnotesize{$y_1$} };
		\draw [->,thin] (i1.north) -- ([yshift=-1pt] o1.south);
		
		\foreach \p/\i/\x/\y in {1/2/$y_2$/\shortpad,2/3/\shortmask/$y_3$,3/4/$y_4$/\shortpad,4/5/\shorteos/\shortpad}
		{
			\node [inputnode, anchor=west] (i\i) at ([xshift=\hseg] i\p.east) {\footnotesize{\x} };					
			\node [outputnode, anchor=south] (o\i) at ([yshift=\vseg] i\i.north) {\footnotesize{\y} };
		
		    \draw [->,thin] (i\i.north) -- ([yshift=-1pt] o\i.south);
		}
		\node[decnode] at ([yshift=0.5\vseg] i3.north) {\footnotesize{Skip-CMLM Decoder}};
		
		\end{scope}			
		\end{tikzpicture}
		}
		\\
			
		\end{tabular}
	\end{center}
	
	\begin{center}
		\vspace{-.5em}
		\caption{Examples of training samples for four tasks, in which (a) and (b) are auxiliary tasks and (c) and (d) are primary tasks. For the sake of clarity, we omit the source sequence. \shortbos/\shorteos/\shortpad/\shortmask represents the special token for \bos/\eos/\pad/\mask, respectively. \spe{B$_2$} is the \bos for k=2. Loss at \spe{P} is ignored. }
	\label{fig:train_sample}
		\vspace{-1.2em}
	\end{center}
\end{figure*}

\noindent
Next, we elaborate on how to train the HRT model efficiently and effectively. Please refer to Appendix~\ref{appendix:training_algo} for the entire training algorithm.

\paragraph{Multi-Task Framework}
We learn HRT through joint training of four tasks, including two primary tasks (\texttt{TASK-SKIP-AT}, \texttt{TASK-SKIP-CMLM}) and two auxiliary tasks (\texttt{TASK-AT}, \texttt{TASK-CMLM}).
All tasks use cross-entropy as the training objective.
Figure~\ref{fig:train_sample} illustrates the differences in training samples among these tasks.
It should be noted that \texttt{TASK-SKIP-AT} shrinks the sequence length from $N$ to $N/k$ compared to \texttt{TASK-AT}, while the token positions follow the original sequence. For example, in Figure~\ref{fig:train_sample} (c), the position of \texttt{TASK-SKIP-AT} input (\spe{B$_2$}, $y_2$, $y_4$) is (0, 2, 4) instead of (0, 1, 2).
Involving auxiliary tasks is necessary because the two primary tasks cannot fully leverage all tokens in the sequence due to the fixed $k$.
For example, in Figure~\ref{fig:train_sample} (c) and (d), $y_1$ and $y_3$ have no chance to be learned as the decoder input of either \texttt{TASK-SKIP-AT} or \texttt{TASK-SKIP-CMLM}. 

\paragraph{Curriculum Learning}
To ensure that the model is not overly biased towards auxiliary tasks, we propose gradually transferring the training tasks from auxiliary tasks to primary tasks through curriculum learning \citep{bengio-cl}.  
More concretely, given a batch of original sentence pairs $\mathcal{B}$ and let the proportion of primary tasks in $\mathcal{B}$ be $p_k$, we start with $p_k$=0 and construct the training samples of \texttt{TASK-AT} and \texttt{TASK-CMLM} for all pairs. 
Then we gradually increase $p_k$ to introduce more learning signals for \texttt{TASK-SKIP-AT} and \texttt{TASK-SKIP-CMLM} until $p_k$=1. In implementation, we schedule $p_k$ by:
\begin{equation}
	\label{eq:skip_rate}
	p_k = ( t/T )^\lambda,
\end{equation}
where $t$ and $T$ are the current and total training steps. $\lambda$ is a hyperparameter, and we use $\lambda$=1 to increase $p_k$ linearly for all experiments.

\begin{table}[t]
\begin{tabular}{cc}
     
     \begin{minipage}[t]{.4\textwidth}
     
	\centering
	\resizebox{\textwidth}{!}
	{
		\begin{tabular}{c c}
				\toprule[1pt]
				\textbf{Method} & \textbf{Generation}   \\ \hline
				SAT & $a,b \rightarrow c,d \rightarrow e,f$ \\
			    RecoverSAT & $a,c,e \rightarrow b,d,f$ \\
				HRT (Our) & $a \rightarrow c \rightarrow e \dashrightarrow b,d,f$ \\
				\bottomrule[1pt]
			\end{tabular}
	}
\vspace{-.5em}
	\captionof{table}{Examples of generating the sequence of ``$a, b, c, d, e, f$'' by different methods. ``$\rightarrow$'' denotes a new decoding step conditioned on the prefix with beam search, while ``$\dashrightarrow$'' is its greedy search version. } 
	\label{table:diff_example}	
	\vspace{-.5em}
     \end{minipage}
     &
    \begin{minipage}[t]{.55\textwidth}
    \resizebox{\textwidth}{!}
    {
        \begin{tabular}{c c c c c c}
				\toprule[1pt]
				\textbf{b$_{at}$} & \textbf{b$_{nat}$} & \textbf{\entoro}  &  \textbf{\entode} & \textbf{$\alpha$(AG)} & \textbf{$\alpha$(AC)}   \\ \hline
				1 & 1 & 34.09 & 28.34 & \textbf{2.1} & \textbf{2.8} \\
				5 & 1 & 34.24 & 28.49 & 1.7 & 1.6 \\
				5 & 5 & \textbf{34.36} & \textbf{28.65} & N/A & 1.1 \\
				\bottomrule[1pt]
		\end{tabular}
	}
	
	\vspace{-.5em}
		\captionof{table}{Effects of different beam sizes in HRT. $\alpha$(AG) and $\alpha$(AC) denotes the average relative speedup ratio with batch size \{1,8,16,32\} on GPU and CPU, respectively (see Appendix~\ref{appendix:variant_speedup} for details). ``N/A'' denotes decoding failed with batch size 32 due to insufficient GPU memory. }
		\label{table:beam_size}	
		\vspace{-.5em}

    \end{minipage}
    
     \\
     
\end{tabular}
\end{table}

\paragraph{Mixed Distillation}

Conventional NAT models use smoother distillation data generated by AT models instead of raw data. 
However, using only distillation data may lose some important information (e.g., rare words) contained in the original data \cite{ding2020understanding}.
To combine the best of both worlds, we propose a simple but effective approach -- Mixed Distillation (MixDistill).
During training, MixDistill randomly samples the target sentence from the raw version $\vy$ with probability $p_{raw}$ or its distillation version $\vy^*$ with probability $1-p_{raw}$, where $p_{raw}$ is a hyperparameter \footnote{Training with only raw data or distillation data can be regarded as the special case of MixDistill as $p_{raw}$=1 or $p_{raw}$=0.}. 
Compared with recent studies, our method is easier to implement: HRT does not rely on external word alignment \cite{ding2020understanding}, and also avoids the time-consuming bi-directional distillation process \cite{ding2021Rejuvenating}. 
MixDistill makes the HRT model less prone to overfitting in simple tasks (e.g., WMT'16 \entoro), but what we emphasize is that the performance of HRT does not depend entirely on it. See Table~\ref{table:distill_strategy} for more details.

\subsection{Inference}
\label{sec:hrt_infer}

HRT adopts two-stage generation strategy. In the first stage, the Skip-AT decoder starts from \spe{BOS$_k$} to autoregressively generate a discontinuous target sequence $\hat{\vy}_{at}=(z_1, z_2, \ldots, z_m)$ with chunk size $k$ until meeting \eos. Then we construct the input of Skip-CMLM decoder $\vy_{nat}$ by appending $k-1$ [\texttt{MASK}]s before every $z_i$. The final translation is generated by replacing all [\texttt{MASK}]s with the predicted tokens by the Skip-CMLM decoder with one iteration. If there are multiple [\texttt{EOS}]s existing, we truncate to the first \eos. Note that the beam size b$_{at}$ in Skip-AT can be different from the beam size b$_{nat}$ in Skip-CMLM, as long as b$_{at}$ $\ge$ b$_{nat}$: We only feed the top b$_{nat}$ Skip-AT hypotheses to Skip-CMLM decoder. In the implementation, we use standard beam search in Skip-AT (b$_{at}>$1) and greedy search in Skip-CMLM (b$_{nat}$=1) because b$_{nat}$=1 is sufficient to obtain competitive BLEU scores, thanks to the good context provided by Skip-AT. 
Table~\ref{table:beam_size} gives a detailed discussion about the beam size setting in HRT.
Finally, we choose the translation hypothesis with the highest score $\rm{S}(\hat{\vy})$ by:
\begin{eqnarray}
	\label{eq:inf_score}
	\underbrace{\sum_{i=1}^{m}\log{P}(z_i|\vx,\vz_{<i})}_{\textrm{Skip-AT score}}  
	+ 
	\underbrace{\sum_{i=0}^{m-1} \sum_{j=1}^{k-1} \log{P}(\hat{y}_{i \times k+j}|\vx,\vy_{nat})}_{\textrm{Skip-CMLM score}} \textrm{, where }  z_i=\hat{y}_{i \times k}.
\end{eqnarray}

\subsection{Discussion}

The basic idea of HRT is to apply AT and NAT in sequence, which has been investigated by \citet{kaiser2018fast,ran2019guiding,akoury2019syntactically}. 
The main difference between these methods is the content of AT output, such as latent variable \citep{kaiser2018fast}, reordered source token \citep{ran2019guiding}, and syntactic label \citep{akoury2019syntactically}. In contrast, our approach uses the deterministic target token as \citet{ghazvininejad-etal-2019-mask}. 
Coupling different decoding paradigms in one process is another line to incorporate AT and NAT. For example, SAT \cite{wang2018semi} embeds chunk-level NAT into the AT process, while RecoverSAT \cite{ran-etal-2020-learning} does the opposite. Table~\ref{table:diff_example} shows the differences between HRT and these two methods.
Although HRT needs more decoding steps, its non-autoregressive process is inexpensive due to the use of greedy search. In contrast, SAT and RecoverSAT require larger beams to explore translations in different lengths.

\begin{table*}[t]
	\begin{center}
		\resizebox{0.95\textwidth}{!}
		{
			\begin{tabular}{l l c c c c c c}
			\toprule[1pt]
			
			\multicolumn{2}{c}{\multirow{2}{*}{\textbf{System}}} & \multicolumn{1}{c}{\multirow{2}{*}{\textbf{Iterations}}} &
		    \multicolumn{2}{c }{\textbf{WMT'16}} & 	\phantom{abc} & 
			\multicolumn{2}{c }{\textbf{WMT'14}} \\
			\cmidrule{4-5} \cmidrule{7-8}
			
			~ & ~ & ~ & \textbf{En-Ro} & \textbf{Ro-En} & &  \textbf{En-De} & \textbf{De-En} \\
			\midrule
			
			\multicolumn{8}{c}{\textit{Existing systems}} \\ \midrule
			\multicolumn{1}{c}{\multirow{5}{*}{\rotatebox{90}{\textbf{NAT}}}} &
			
			FCL-NAT  \citep{guo2019fine} & 1 & - & - && 25.75 & 29.50 \\ 
			~ & FlowSeq  \cite{ma2019flowseq} & 1  & 32.20 & 32.84 & &  25.31 & 30.68 \\ 
			~ & AXE \cite{ghazvininejad2020aligned}  & 1 & 30.75 & 31.54 && 23.53 & 27.90 \\
			~ & GLAT \cite{qian-etal-2021-glancing} & 1 & 32.87 & 33.84 && 26.55 & 31.02  \\
			~ & Fully-NAT \cite{gu2021fully} & 1 &  33.79 & 34.16 && 27.49 & 31.39  \\
			
			\cdashline{1-8}
			
			\multicolumn{1}{c}{\multirow{8}{*}{\rotatebox{90}{\textbf{Iterative NAT}}}} &
			 CMLM \citep{ghazvininejad-etal-2019-mask} & 10 & 33.08 & 33.31 && 27.03 & 30.53 \\ 
			~ & LevT  \citep{gu2019levenshtein} & Adaptive & - & - && 27.27 & -  \\ 
			~ & JM-NAT  \citep{guo2020jointly}  & 10 & 33.52 & 33.72 && 27.69 & 32.24  \\
			~ & SMART \cite{ghazvininejad2020semi} & 10 & - & - && 27.65 & 31.27 \\
			~ & DisCO \cite{kasai2020non}  & Adaptive & 33.22 & 33.25 && 27.34 & 31.31 \\
			~ & Imputer \cite{saharia2020non} & 8 & 34.40 & 34.10 && 28.20 & 31.80 \\
			~ & RewriteNAT \cite{geng-etal-2021-learning} & Adaptive & 33.63 & 34.09 && 27.83 & 31.52 \\
			~ & CMLMC \citep{huang2022improving}  & 10 &  34.57 & 34.13 && 28.37 & 31.41 \\
			\cdashline{1-8}
			
			\multicolumn{1}{c}{\multirow{4}{*}{\rotatebox{90}{\textbf{Semi-NAT}}}} &
			SAT \cite{wang2018semi}  & $N/2$ & - & - && 26.90 & - \\ 
			~ & SynST \cite{akoury2019syntactically}  & $N/6+1$  & - & - &&  20.74$^\dag$ & 25.50$^\dag$ \\ 
			
			~ & ReorderNAT \citep{ran2019guiding} & $N+1$ & 31.70 & 31.99 && 26.49 & 31.13 \\
			~ & RecoverSAT \citep{ran-etal-2020-learning} & $N/2$ &  32.92 & 33.19 && 27.11 &  31.67 \\
		
			\midrule
			
			\multicolumn{8}{c}{\textit{Our implementations}} \\ \midrule
			
			\multicolumn{1}{c}{\multirow{5}{*}{}} &
			AT + Raw  & $N$ & 34.25(34.2$^\dag$) & 34.40(34.0$^\dag$) && 27.45(26.9$^\dag$) & 31.86(31.6$^\dag$) \\
			~ & AT-20L + Raw (teacher for \enbothde) & $N$ & - & - && 28.79(28.2$^\dag$) & 33.02(32.8$^\dag$) \\
			~ & SAT + MixDistill & $N/2$ & - & - && 26.67(26.2$^\dag$) & - \\
			~ & HRT + MixDistill & $N/2 + 1$ & 34.24(34.2$^\dag$) & 34.35(34.0$^\dag$) &~& 28.49(27.9$^\dag$) & 32.28(32.0$^\dag$)  \\
			~ & HRT-20L + Mixstill & $N/2 + 1$ & - & - && \textbf{29.14(28.6$^\dag$)} & \textbf{33.21(32.9$^\dag$)}  \\

			\bottomrule[1pt]
			\end{tabular}
		}
		\vspace{-.5em}
		\caption{The BLEU \& SacreBLEU (denoted by $^\dag$) scores of our proposed HRT and the baselines on four WMT tasks. Unless otherwise stated, we use a beam size of 5. ``Adaptive'' denotes dynamic iterations. ``20L'' stands for using a 20-layer encoder. The teacher models of \entoro and \rotoen have BLEU scores of 34.28 and 33.99, respectively, obtained from \citet{ghazvininejad-etal-2019-mask}.
		}
		\label{table:main_results}
		\vspace{-.5em}
	\end{center}
\end{table*}

\section{Experimental Results}
\label{sec:exp}

\paragraph{Setup}
We mainly conduct experiments on four widely used WMT tasks: WMT'16 English$\leftrightarrow$Romanian (\enbothro, 610k) and WMT'14 English$\leftrightarrow$German (\enbothde, 4.5M). We replicate the same data processing for fair comparisons as \citet{ghazvininejad-etal-2019-mask}. To verify the effectiveness in long-distance language pairs, we also test it in the NIST Chinese-English (\zhtoen, 1.8M) translation task following the setup of \citet{wang2018multi}.
We use the \enbothro distillation data published by \citet{ghazvininejad-etal-2019-mask}. However, for \enbothde and \zhtoen, we retrained the AT teacher models because \citet{ghazvininejad-etal-2019-mask} did not release the corresponding distillation data. Specifically, we use the deep PreNorm Transformer-Base with a 20-layer encoder and the standard Transformer-Base as teacher models for \enbothde and \zhtoen, respectively.
We run all experiments on four 2080Ti GPUs. 
Unless noted otherwise, we use the chunk size $k$=2. We set p$_{raw}$=0.5 for \enbothro and \zhtoen, p$_{raw}$=0.2 for \enbothde according to validation sets. 
We fine-tune HRT models on pre-trained AT models and take 100k/300k/100k training steps for \enbothro/\enbothde/\zhtoen, respectively. 
Other training hyperparameters are the same as \citet{vaswani2017attention} or \citet{wang2019learning} (deep-encoder). We report both case-sensitive tokenized BLEU scores and SacreBLEU \footnote{Signature: BLEU+case.mixed+lang.\textit{source}-\textit{target}+numrefs.1+smooth.exp+tok.13a+version.1.5.1}.

\paragraph{Beam Size on HRT} We first verify the influence of two beam sizes of HRT (b$_{at}$ and b$_{nat}$) on the BLEU score and relative acceleration ratio. We test three different setups, and the results are listed in Table~\ref{table:beam_size}. Consistent with our observations in synthetic experiments, using b$_{nat}$=1 only slightly reduces BLEU but significantly improves decoding efficiency. Considering the trade-off between translation quality and speed and fair comparison with baselines \footnote{Most prior related work uses beam size of 5.}, we use b$_{at}$=5 and b$_{nat}$=1 unless otherwise stated.

\paragraph{Main Results}

\begin{table}[t]
\begin{tabular}{cc}
    \begin{minipage}[t]{.45\textwidth}
    \centering
    \resizebox{\textwidth}{!}
    {
        	\begin{tabular}{c c c c}
				\toprule[1pt]
				\textbf{Model} & \textbf{MT04} & \textbf{MT05} & \textbf{MT08}   \\ \hline
				AT (teacher) & 43.86 &	52.91	& 33.94 \\
				CMLM$_{10}$	& 42.47 &	52.16 &	33.09 \\
				HRT	& \textbf{43.93} &	\textbf{53.02} &	\textbf{34.33} \\
				\bottomrule[1pt]
			\end{tabular}
	}
	\vspace{-.5em}
		\captionof{table}{BLEU scores on NIST \zhtoen task.}
		\label{table:trans_zh2en}	
		\vspace{-.5em}
	
    \end{minipage}
     &  
     \begin{minipage}[t]{.45\textwidth}
     
	\centering
	\resizebox{\textwidth}{!}
	{
	    \begin{tabular}{c c c c c}
			\toprule[1pt]
			
			\multicolumn{1}{c}{\textbf{Chunk}} &
			\multicolumn{1}{c}{\textbf{Valid}} &
			\multicolumn{1}{c}{\textbf{Test}} &
			\multicolumn{1}{c}{\textbf{$\alpha$(AG)}} &
			\multicolumn{1}{c}{\textbf{$\alpha$(AC)}} 
			\\ \midrule
			
			2 & \textbf{26.45}  & \textbf{28.49} & 1.7 & 1.6 \\
			3 & 26.22  & 27.98 & 2.5 & 2.3 \\
			4 & 25.56  & 27.17 & \textbf{3.2} & \textbf{2.9} \\	 
			\bottomrule[1pt]
		\end{tabular}
	}
	\vspace{-.5em}
	\caption{Effects of chunk size ($k$) on BLEU score and $\alpha$. 
	}
	\label{table:chunk_k}
	\vspace{-.5em}
	
     \end{minipage}
     \\
     
\end{tabular}
\end{table}

Table~\ref{table:main_results} reports the BLEU and SacreBLEU scores on four WMT tasks. 
Our HRT outperforms most existing NAT, IR-NAT, and Semi-NAT models and establishes new state-of-the-art results on \entode (See Appendix~\ref{appendix:case_study} for case studies.). 
Compared to our re-implemented SAT with the same MixDistill, HRT obtains an improvement of +2.0 BLEU points. We compared different data strategies in more detail in Section~\ref{sec:analysis}.
Besides, in line with \citet{guo2020jointly}, when using a deeper encoder, HRT-20L can further improve by approximately +0.5 BLEU. 
We highlight that HRT can achieve equivalent or even better performance than the teacher model when having the same model capacity.
In Table~\ref{table:trans_zh2en}, we also report the experimental results on the \zhtoen task. HRT is once again superior to the original AT and CMLM model, which indicates that the effectiveness of HRT is agnostic to language pairs.

\begin{table}[t]
\begin{tabular}{cc}
    \begin{minipage}[t]{.5\textwidth}
    \resizebox{\textwidth}{!}
    {
    \begin{tabular}{l c c c c c c c}
		\toprule[1pt]
		
		\multicolumn{1}{l}{\multirow{2}{*}{\textbf{System}}} & \multicolumn{3}{c}{\multirow{1}{*}{\textbf{\entoro}}} &
		\phantom{a} &
		\multicolumn{3}{c}{\multirow{1}{*}{\textbf{\entode}}} \\
		\cmidrule{2-4} \cmidrule{6-8}
		
		~ & \textbf{R} & \textbf{D} & \textbf{MD} && \textbf{R} & \textbf{D} & \textbf{MD}   \\ \midrule
		AT & \textbf{34.25}  & 33.19 & 33.92 && \textbf{27.45} & 28.24 & 28.14 \\ 
		SAT & - & - & - && 22.07 & 26.45 & 26.67 \\
		HRT & 33.59 & \textbf{33.20} & \textbf{34.24} && 26.69 & \textbf{28.30} & \textbf{28.49} \\

		\bottomrule[1pt]
	\end{tabular}
	}
	
	\vspace{-.5em}
	\caption{The BLEU scores against different data strategies (R: Raw, D: Distill, MD: MixDistill) 
	. }
	\label{table:distill_strategy}
	\vspace{-.5em}

    \end{minipage}
     &  
     \begin{minipage}[t]{.4\textwidth}
    \resizebox{\textwidth}{!}
    {
        \begin{tabular}{c c c c}
			\toprule[1pt]
			
			\textbf{Skip-AT} & \textbf{Skip-CMLM} & \textbf{BLEU} & \textbf{$\Delta$} \\
			\midrule
			
			HRT & HRT & 28.49 & ref. \\
			HRT-20L & HRT-20L & 29.14 & +0.65 \\
			HRT & HRT-20L & 28.58 & +0.09 \\
			HRT-20L & HRT & 29.03 & +0.54 \\
			
			\bottomrule[1pt]
			\end{tabular}
	}
	\vspace{-.5em}
	\caption{Swapping Skip-AT and Skip-CMLM between HRT and HRT-20L.}
	\label{table:importance_at_nat}
	\vspace{-.5em}
    \end{minipage}
     \\
     
\end{tabular}
\end{table}

\section{Analysis}
\label{sec:analysis}

\paragraph{Impact of Chunk Size}
We tested chunk size $k$ on the \entode task as shown in Table~\ref{table:chunk_k}, where we can see that:
(1) A large $k$ has a more significant speedup on the GPU because fewer autoregressive steps are required;
(2) As $k$ increases, the performance of HRT drops sharply. For example, $k$=4 is about 1.32 BLEU points lower than $k$=2 on the test set.
It indicates that the training difficulty of Skip-AT increases as $k$ becomes larger. We think that skip-generation may require more fancy training algorithms, which is left for our future work.

\paragraph{Data Strategy}
As shown in Table~\ref{table:distill_strategy}, we compared three data strategies, including raw data (\texttt{R}), sequence-level knowledge distillation (\texttt{D}), and proposed mixed distillation (\texttt{MD}).
Overall, \texttt{MD} is superior to other methods across the board, indicating that training with raw and distillation data is complementary. We emphasize that HRT's strong performance does not mainly come from \texttt{MD}. For example, even if using \texttt{D}, HRT can achieve excellent results on the \entode task. More surprisingly, we found that HRT can be slightly better than AT models in the same network scale trained by \texttt{D}. We attribute it to two reasons: (1) HRT is fine-tuned on a well-trained AT model; (2) Multi-task learning on autoregressive and non-autoregressive tasks has better regularization than training alone.

\paragraph{Importance of Skip-AT \& Skip-CMLM}
We try to understand the importance of Skip-AT and Skip-CMLM for HRT. To this end, we swap the intermediate results of two different HRT models (refer to Model \texttt{A} and \texttt{B}, respectively). Specifically, we first use the Skip-AT decoder of \texttt{A} to generate its discontinuous target sequence. Then we let \texttt{B}'s Skip-AT decoder force decoding this sequence to obtain corresponding encoding representations and autoregressive model scores. Finally, \texttt{B} uses its Skip-CMLM decoder to generate the complete translation result according to them. The order of \texttt{A} and \texttt{B} can be exchanged. In practice, we use two models (HRT and HRT-20L) with large performance gap as \texttt{A} and \texttt{B} respectively. As shown in Table~\ref{table:importance_at_nat}, we find that using the strong model's Skip-AT brings more improvement (+0.54 BLEU) than using its Skip-CMLM (+0.09 BLEU). This result aligns with our claim that a good partial target context is critical.

\paragraph{Deep-encoder-shallow-decoder Architecture}

\citet{kasai2020deep} point out that AT with deep-encoder-shallow-decoder architecture can substantially speed up without losing translation accuracy.
We also compare HRT and AT in this setting: 12-layer encoder and 1-layer decoder, denoted by HRT$_{12-1}$ and AT$_{12-1}$, respectively. 
AT$_{12-1}$ and HRT$_{12-1}$ use the same mixed distillation data for a fair comparison. As can be seen from Table~\ref{table:deep_shallow}, both AT$_{12-1}$ and HRT$_{12-1}$ benefit from the change of layer allocation, gaining comparable BLEU scores and double decoding speed as the vanilla models. Specifically, HRT$_{12-1}$ achieves an average acceleration of 4.2x/3.1x than AT baselines. \citet{kasai2020deep} report that CMLM does not enjoy this architecture, which indicates the success of HRT$_{12-1}$ comes from using Skip-AT instead of Skip-CMLM.

\begin{table}[t]
\begin{tabular}{M{.63\textwidth}M{.32\textwidth}}
     \definecolor{coquelicot}{rgb}{1.0, 0.22, 0.0}
    \newcommand\satcolor{coquelicot}
    \definecolor{brightpink}{rgb}{1.0, 0.0, 0.5}
    \newcommand\dehrtcolor{brightpink}
    \definecolor{airforceblue}{rgb}{0.36, 0.54, 0.66}
    \newcommand\hrtcolor{airforceblue}
	\renewcommand\arraystretch{0.0}
	\setlength{\tabcolsep}{0pt}
    \begin{minipage}[t]{.63\textwidth}
    \resizebox{.9\textwidth}{!}
    {
        \newlength{\marksize}
        \setlength{\marksize}{5pt}
		\begin{tikzpicture}
			\begin{axis}[domain  = 0.97:2.3,
				width=\textwidth,
				height=.53\textwidth,
				samples = 100,
				ymin    = 17,
				ymax    = 23,
				xmin    = -0.5,
				xmax    = 12.5,
				separate axis lines,
                x axis line style= { draw opacity=0 },
                y axis line style= { draw opacity=0 },
				yticklabels={18,25,26,27,28,29},
				ytick={18,19,20,21,22,23},
				xtick={0,4,8,9,10,11,12},
				xticklabels={0,1,2,3,4,5,10},
				tick label style={font=\scriptsize},
				xlabel near ticks,
				ylabel near ticks,
				grid=major,
				legend style={at={(0.75,+1.05)},
					anchor=south,legend columns=6},
				]
			
        \tikzstyle{gpunode1} = [circle,minimum size=\marksize, inner sep=0pt,draw=black!80]
        \tikzstyle{gpunode8} = [rectangle,minimum size=\marksize, inner sep=0pt,draw=black!80]
        \tikzstyle{gpunode16} = [diamond,minimum size=\marksize+1, inner sep=0pt,draw=black!80]
        \tikzstyle{gpunode32} = [star,minimum size=\marksize+1, inner sep=0pt,draw=black!80]
        \tikzstyle{cpunode1} = [circle,minimum size=\marksize, inner sep=0pt]
        \tikzstyle{cpunode8} = [rectangle,minimum size=\marksize, inner sep=0pt]
        \tikzstyle{cpunode16} = [diamond,minimum size=\marksize+1, inner sep=0pt]
        \tikzstyle{cpunode32} = [star,minimum size=\marksize+1, inner sep=0pt]
        
        \newcommand\cmlmonecolor{darkblue}
        \newcommand\cmlmonebleu{18.05} 
        \node[gpunode1, fill=\cmlmonecolor] at (axis cs:11.92,\cmlmonebleu){};
        \node[gpunode8, fill=\cmlmonecolor] at (axis cs:10.8,\cmlmonebleu){};
        \node[gpunode16, fill=\cmlmonecolor] at (axis cs:8.8,\cmlmonebleu){};
        \node[gpunode32, fill=\cmlmonecolor] at (axis cs:6,\cmlmonebleu){};
        
        \node[cpunode1,draw=\cmlmonecolor] at (axis cs:11.08,\cmlmonebleu){};
        \node[cpunode8,draw=\cmlmonecolor] at (axis cs:8.7,\cmlmonebleu){};
        \node[cpunode16,draw=\cmlmonecolor]  at (axis cs:8.5,\cmlmonebleu){};
        \node[cpunode32,draw=\cmlmonecolor]  at (axis cs:8.3,\cmlmonebleu){};
       
        \newcommand\cmlmfourcolor{ugreen}
        \newcommand\cmlmfourbleu{19.94} 
        
        \node[gpunode1, fill=\cmlmfourcolor]  at (axis cs:9.8,\cmlmfourbleu){};
        \node[gpunode8, fill=\cmlmfourcolor]  at (axis cs:8.3,\cmlmfourbleu){};
        \node[gpunode16, fill=\cmlmfourcolor]  at (axis cs:5.6,\cmlmfourbleu){};
        \node[gpunode32, fill=\cmlmfourcolor]  at (axis cs:3.2,\cmlmfourbleu){};
        
        \node[cpunode1,draw=\cmlmfourcolor] at (axis cs:7.04,\cmlmfourbleu){};
        \node[cpunode8,draw=\cmlmfourcolor]  at (axis cs:3.68,\cmlmfourbleu){};
        \node[cpunode16,draw=\cmlmfourcolor]  at (axis cs:3,\cmlmfourbleu){};
        \node[cpunode32,draw=\cmlmfourcolor]  at (axis cs:2.52,\cmlmfourbleu){};
        
        \newcommand\nodetag{cmlm10g}
        \newcommand\nodecolor{lyellow}
        \newcommand\bleu{21.03} 
        \node[gpunode1, fill=\nodecolor] (\nodetag1) at (axis cs:6.96,\bleu){};
        \node[gpunode8, fill=\nodecolor] (\nodetag8) at (axis cs:4.76,\bleu){};
        \node[gpunode16, fill=\nodecolor] (\nodetag16) at (axis cs:2.76,\bleu){};
        \node[gpunode32, fill=\nodecolor] (\nodetag32) at (axis cs:1.52,\bleu){};
        
        \node[cpunode1,draw=\nodecolor] (\nodetag1) at (axis cs:3.2,\bleu){};
        \node[cpunode8,draw=\nodecolor] (\nodetag8) at (axis cs:1.52,\bleu){};
        \node[cpunode16,draw=\nodecolor] (\nodetag16) at (axis cs:1.24,\bleu){};
        \node[cpunode32,draw=\nodecolor] (\nodetag32) at (axis cs:1.04,\bleu){};
        
        \newcommand\hrttag{hrtg}
        \newcommand\hrtbleu{22.49} 
        \node[gpunode1, fill=\hrtcolor] (\hrttag) at (axis cs:7.16,\hrtbleu){};
        \node[gpunode8, fill=\hrtcolor] (\hrttag) at (axis cs:7.08,\hrtbleu){};
        \node[gpunode16, fill=\hrtcolor] (\hrttag) at (axis cs:6.96,\hrtbleu){};
        \node[gpunode32, fill=\hrtcolor] (\hrttag) at (axis cs:6.6,\hrtbleu){};
        
        \node[cpunode1,draw=\hrtcolor] (\hrttag) at (axis cs:6.76,\hrtbleu){};
        \node[cpunode8,draw=\hrtcolor] (\hrttag) at (axis cs:6.76,\hrtbleu){};
        \node[cpunode16,draw=\hrtcolor] (\hrttag) at (axis cs:6.24,\hrtbleu){};
        \node[cpunode32,draw=\hrtcolor] (\hrttag) at (axis cs:5.84,\hrtbleu){};
        
        \newcommand\atg{atg}
        \newcommand\atcolor{mypurple}
        \newcommand\atbleu{22.4} 
        \node[gpunode1, fill=\atcolor] (\atg1) at (axis cs:8.91,\atbleu){};
        \node[gpunode8, fill=\atcolor] (\atg8) at (axis cs:8.81,\atbleu){};
        \node[gpunode16, fill=\atcolor] (\atg16) at (axis cs:8.6,\atbleu){};
        \node[gpunode32, fill=\atcolor] (\atg32) at (axis cs:8.4,\atbleu){};
        
        \node[cpunode1,draw=\atcolor] (\atg1) at (axis cs:8.76,\atbleu){};
        \node[cpunode8,draw=\atcolor] (\atg8) at (axis cs:8.11,\atbleu){};
        \node[cpunode16,draw=\atcolor] (\atg16) at (axis cs:7.52,\atbleu){};
        \node[cpunode32,draw=\atcolor] (\atg32) at (axis cs:7.28,\atbleu){};
       
         
        \newcommand\dehrtbleu{22.44} 
        \node[gpunode1, fill=\dehrtcolor] at (axis cs:10.64,\dehrtbleu){};
        \node[gpunode8, fill=\dehrtcolor] at (axis cs:10.32,\dehrtbleu){};
        \node[gpunode16, fill=\dehrtcolor] at (axis cs:9.99,\dehrtbleu){};
        \node[gpunode32, fill=\dehrtcolor] at (axis cs:9.68,\dehrtbleu){};
        
        \node[cpunode1,draw=\dehrtcolor]  at (axis cs:8.76,\dehrtbleu){};
        \node[cpunode8,draw=\dehrtcolor]  at (axis cs:8.11,\dehrtbleu){};
        \node[cpunode16,draw=\dehrtcolor]  at (axis cs:7.52,\dehrtbleu){};
        \node[cpunode32,draw=\dehrtcolor]  at (axis cs:7.28,\dehrtbleu){};
        
        
        \newcommand\satbleu{20.45} 
        \node[gpunode1, fill=\satcolor] at (axis cs:6.56,\satbleu){};
        \node[gpunode8, fill=\satcolor] at (axis cs:6.32,\satbleu){};
        \node[gpunode16, fill=\satcolor] at (axis cs:6.08,\satbleu){};
        \node[gpunode32, fill=\satcolor] at (axis cs:6.12,\satbleu){};
        
        \node[cpunode1,draw=\satcolor]  at (axis cs:6.08,\satbleu){};
        \node[cpunode8,draw=\satcolor]  at (axis cs:5.04,\satbleu){};
        \node[cpunode16,draw=\satcolor]  at (axis cs:4.8,\satbleu){};
        \node[cpunode32,draw=\satcolor]  at (axis cs:5.08,\satbleu){};
        
        \newcommand\baselinebleu{21.45} 
        \draw [dashed, line width=0.5pt] (-10,\baselinebleu) -- (20,\baselinebleu);
        \draw [dashed, line width=0.5pt] (4,0) -- (4,30);
        \node [star,star points=10, fill=red, inner sep=0pt, minimum size=10pt] at (axis cs: 4,\baselinebleu){};
        
\path[-] (rel axis cs:0,0)     coordinate(leftstart)
          --(rel axis cs:0,0.2) coordinate(interruptleftA)
         (rel axis cs:0,0.3)  coordinate(interruptleftB)
         --(rel axis cs:0,1)   coordinate(leftstop);

\path[-] (rel axis cs:0,1)     coordinate(topstart)
          --(rel axis cs:1,1)   coordinate(topstop);

\path[-] (rel axis cs:0,0)     coordinate(botstart)
         --(rel axis cs:0.91,0) coordinate(interruptbotA)
         (rel axis cs:0.95,0)  coordinate(interruptbotB)
         --(rel axis cs:1,0)   coordinate(botstop);

\path[-] (rel axis cs:1,0)     coordinate(rightstart)
          --(rel axis cs:1,1)   coordinate(rightstop);

			\end{axis}
\draw(leftstart)-- (interruptleftA) decorate[decoration=zigzag]{--(interruptleftB)} -- (leftstop);

\draw(botstart)-- (interruptbotA) decorate[decoration=zigzag]{--(interruptbotB)} -- (botstop);

\draw(topstart)-- (topstop);
\draw(rightstart)-- (rightstop);

\scriptsize
		\node (l0) at (0.15,0.35) {};
		\draw[darkblue, line width=2pt] (0.15,1.15) -- plot[](0.35,1.15) -- (0.6,1.15) node[right] (l1) {CMLM$_1$};
		\draw[mypurple, line width=2pt] (1.8,1.15) -- plot[](2.05,1.15) -- (2.3,1.15) node[right] (l2) {AT$_{12-1}$};
		
		\draw[ugreen, line width=2pt] (0.15,0.85) -- plot[](0.35,0.85) -- (0.6,0.85) node[right] (l3) {CMLM$_{4}$};
		\draw[\satcolor, line width=2pt] (1.8,0.85) -- plot[](2.05,0.85) -- (2.3,0.85) node[right] (l4) {SAT};
		\draw[lyellow, line width=2pt] (0.15,0.55) -- plot[](0.35,0.55) -- (0.6,0.55) node[right] (l5) {CMLM$_{10}$};
		\draw[\hrtcolor, line width=2pt] (1.8,0.55) -- plot[](2.05,0.55) -- (2.3,0.55) node[right] (l6) {HRT};
		\draw[\dehrtcolor, line width=2pt] (0.15,0.3) -- plot[](0.35,0.3) -- (0.6,0.3) node[right] (l7) {HRT$_{12-1}$};
		\node [star,star points=10, fill=red, inner sep=0pt, minimum size=8pt] at (2.05,0.3) {};
		\draw[red, line width=2pt]  (2.3,0.3) node[right] (l8) {AT};
		
		\begin{pgfonlayer}{background}
		\node[rectangle,draw,inner sep=0pt] [fit = (l0) (l1) (l2) (l3) (l4) (l5) (l6) (l7) (l8)] {};
		\end{pgfonlayer}
		\end{tikzpicture}
	}
	\vspace{-2em}
     \end{minipage}
     &
    \begin{minipage}[t]{.32\textwidth}
    
	
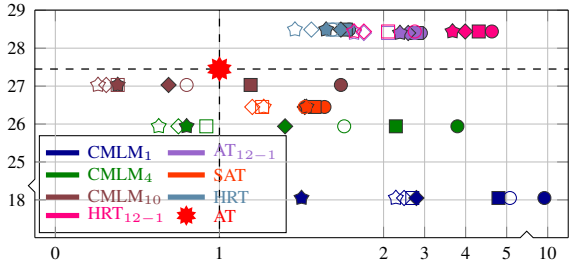
\captionof{figure}{Comparing relative speedup ratio (x-axis) and BLEU score (y-axis) against decoding batch sizes (1: \ding{108}, 8: $\blacksquare$, 16: $\blacklozenge$, 32: $\bigstar$) and running devices (solid/hollow shapes represent GPU/CPU, respectively). 
	}
	\label{fig:main_speed}
	\vspace{-2em}
    \end{minipage}
    
     \\
     
\end{tabular}
\end{table}

\begin{table}[t]
\begin{tabular}{M{0.55\textwidth}M{.4\textwidth}}
    \begin{minipage}[t]{.55\textwidth}
     
	\centering
	\resizebox{\textwidth}{!}
	{
	\begin{tabular}{l c c c c }
			\toprule[1pt]
			
			\textbf{Model} & \textbf{MD} & \textbf{BLEU} &
			\textbf{$\alpha$(AG)} &
			\textbf{$\alpha$(AC)} \\
			\midrule
			
			AT-20L (teacher) & \ding{55} & 28.87 & - & - \\
			
			AT & \ding{51} & 28.14 &
			ref. &
			ref. \\
			
			AT$_{12-1}$& \ding{51} & 28.18 &
		    2.7 &
			2.1 \\
			
			HRT & \ding{51} & \textbf{28.49} &
			1.7 &
			1.6 \\
			
			HRT$_{12-1}$ & \ding{51} & 28.44 &
			\textbf{4.2} &
			\textbf{3.1} \\
				
			\bottomrule[1pt]
			\end{tabular}
	}
	\vspace{-.5em}
	\caption{Effects of deep-encoder-shallow-decoder architecture on \entode test set.\\}
	\label{table:deep_shallow}
	\vspace{-2.5em}
     \end{minipage}
     &  
     \begin{minipage}[t]{.4\textwidth}
     
	\centering
	\resizebox{.85\textwidth}{!}
	{

	   \begin{tabular}{l c c}
				\toprule[1pt]
				
				\textbf{System} &  \textbf{BLEU} & \textbf{$\Delta$} \\ \midrule
				
				
				HRT ($T$=300k) & 28.49 & ref. \\
				\quad $-$FT & 28.21 & -0.28 \\
				\quad $-$MD (p$_{raw}$=0)  & 28.30 & -0.19 \\
				\quad $-$CL ($p_k$=1) & 27.72 & -0.77 \\
				\quad $-$TS ($T$=100k) & 27.91 & -0.58 \\
				\quad $-$ALL & 26.96 & -1.53 \\
				\bottomrule[1pt]
			\end{tabular}
	}
	\vspace{-.5em}
	\caption{Ablation study on \entode task. }
	\label{table:ablation}
	\vspace{-2.5em}

	 \end{minipage}
     \\
     
\end{tabular}
\end{table}




\paragraph{Ablation Study} 
In Table~\ref{table:ablation}, we conduct ablation studies on the \entode task, including fine-tuning from pre-trained AT (FT), mixed distillation (MD), training steps (TS), and curriculum learning (CL). CL is equivalent to removing auxiliary tasks. All components contribute to the performance, but CL (-0.77) and TS (0.58) are the most critical. 
We also try to exclude them all from the vanilla HRT (-ALL), resulting in a total reduction of 1.53 BLEU points.

\paragraph{Visually Comprehensive Comparison}
We draw Figure~\ref{fig:main_speed} to visually compare the trade-off between BLEU and inference speed of different methods, including CMLM$_1$, CMLM$_4$, CMLM$_{10}$, SAT, AT, AT$_{12-1}$, HRT and HRT$_{12-1}$.
The desired method should have comparable BLEU and faster speed than vanilla AT across different running conditions. 
From this perspective, it can be seen that the CMLM family and SAT are unsatisfactory because of low BLEU or unstable decoding speed. In contrast, AT$_{12-1}$ and our HRT/HRT$_{12-1}$ have higher BLEU and robust acceleration.

\section{Conclusion}
\label{sec:conclusion}
We pointed out that IR-NAT suffers from inference acceleration robustness problems. Inspired by our findings in synthetic experiments, we proposed HRT to combine the advantages of AT and NAT. Experimental results show that our approach outperforms existing Semi-AT methods and promises to be a good substitute for AT due to competitive performance and stable speedup.

\bibliography{custom}
\bibliographystyle{iclr2023_conference}

\appendix

\section{Detailed inference speed}
\label{appendix:variant_speedup}

In Table~\ref{table:speedup_wrt_model_varant}, we list the exact decoding time and relative speedup ratio of different models under varying environments on the \entode test set. When changing the batch size from 1 to 32, the decoding time of AT reduces 20.4x/4.6x on GPU/CPU, respectively, while that of CMLM$_{10}$ only reduces 3.7/0.8x. In contrast, HRT inherits the good character of AT and achieves 18.7x/3.8x speedup.
On the other hand, HRT has more robust acceleration than multi-shot NAT, such as CMLM$_{4}$, CMLM$_{10}$. When using the deep-encoder-shallow-decoder architecture, the performance of HRT$_{12-1}$ approaches the one-shot NAT (CMLM$_1$) on both GPU and CPU. Besides, the overall results of HRT-20L is similar to those of HRT because the translation time is mainly consumed in the decoder. We also report the change of inference speed along with chunk size $k$. 

\begin{table*}[h]
	
	\centering
	\resizebox{\textwidth}{!}
	{
		\begin{tabular}{l ccc ccc ccc ccc c}
			\toprule[1pt]
			
			\multicolumn{1}{l}{\multirow{2}{*}{\textbf{Model}}} &
			\multicolumn{2}{c}{\textbf{$B$=1}} & \phantom{ab} &
			\multicolumn{2}{c}{\textbf{$B$=8}} & \phantom{ab} &
			\multicolumn{2}{c}{\textbf{$B$=16}} & \phantom{ab} &
			\multicolumn{2}{c}{\textbf{$B$=32}} & \phantom{ab} &
			\multicolumn{1}{c}{\textbf{Avg}} \\
			\cmidrule{2-3} \cmidrule{5-6} \cmidrule{8-9} \cmidrule{11-12} \cmidrule{14-14}
			
			~ & \textbf{Time$\downarrow$} & \textbf{$\alpha \uparrow$} && \textbf{Time$\downarrow$} & \textbf{$\alpha \uparrow$} && \textbf{Time$\downarrow$} & \textbf{$\alpha \uparrow$} && \textbf{Time$\downarrow$} & \textbf{$\alpha \uparrow$} &&  \textbf{$\alpha \uparrow$} \\
			\midrule
						
			\multicolumn{14}{c }{\textit{On GPU}} \\ \hline
			AT & 857.2 & 1.0 &&	137.8 & 1.0 && 73.1 & 1.0 && 40.1 & 1.0 &&  1.0 \\
			AT$_{12-1}$ & 	294.7 & 2.9 &&	49.1 & 2.8 && 28.1 & 2.6	&& 16.7 & 2.4 &&   2.7 \\
			CMLM$_{1}$ & \textbf{89.4} & \textbf{9.6} &&	\textbf{28.8} & \textbf{4.8} && 26.3 & 2.8	&& 26.2 & 1.5 &&   \textbf{4.7} \\
			
			CMLM$_{4}$ & 223.5 & 3.8 &&	59.2 & 2.3 && 52.0 & 1.4 && 52.4 & 0.8 &&  2.1 \\
			CMLM$_{10}$ & 492.7 & 1.7 && 116.0 & 1.2 && 106.1 & 0.7	&& 105.0 & 0.4 &&  1.0 \\
			SAT & 523.0 & 1.6 &&	87.1 & 1.6 && 48.0 & 1.5	&& 26.2 & 1.5 &&  1.6 \\
				
			HRT (b$_{at}$=1, b$_{nat}$=1) & 377.5 & 2.3 && 66.4 & 2.1 && 34.9 & 2.1 && 20.5 & 2.0 &&  2.1 \\
			HRT  & 478.9 & 1.8 &&		77.8 & 1.8 && 41.9 & 1.7 && 24.3 & 1.7 &&  1.7 \\
			HRT (b$_{at}$=5, b$_{nat}$=5) & 482.4 & 1.8 &&	81.2 & 1.7 && 46.5 & 1.6 && N/A & N/A && N/A \\
			HRT$_{12-1}$ & 192.5 & 4.6 &&	31.4 & 4.3 && \textbf{18.4} & \textbf{4.0}	&& \textbf{11.1}  & \textbf{3.7} &&  4.2 \\
			HRT (k=3) & 323.9 & 2.6 && 54.9 & 2.5 && 29.5 & 2.5 && 18.2  & 2.2 &&  2.5 \\
			HRT (k=4) & 256.0 & 3.3 && 43.1 & 3.2 && 23.3 & 3.1	&& 12.7  & 3.2 &&  3.2 \\

			\midrule
			
			\multicolumn{14}{c }{\textit{On CPU}} \\ \hline
			AT & 1118.0 & 1.0 &&	314.1 & 1.0	&& 246.3 & 1.0 &&	201.3 & 1.0 &&  1.0\\
			AT$_{12-1}$	&	405.4 & 2.8 &&	149.0	& 2.1 && 130.4 & 1.9	&& 110.7 & 1.8 &&  2.1 \\
		    
		    CMLM$_{1}$ & \textbf{207.3} & \textbf{5.4} &&	116.0 & 2.7 && 97.6 & 2.5	&& 85.9 & 2.3 &&   \textbf{3.2} \\
			CMLM$_{4}$ & 635.1 & 1.8 &&	341.7 & 0.9 && 329.8 & 0.7	&& 319.4 & 0.6 && 1.0 \\
			CMLM$_{10}$ & 1390.9 & 0.8 && 820.1 & 0.4 && 789.3 & 0.3	&& 776.9 & 0.3 && 0.4 \\
			
			SAT & 737.5 & 1.5 &&	248.7 & 1.3 && 205.6 & 1.2	&& 158.9 & 1.3 && 1.3 \\

			HRT (b$_{at}$=1, b$_{nat}$=1) & 457.1 & 2.4 && 116.1 & 2.7 && \textbf{82.4} & \textbf{3.0}  && \textbf{65.9}  & \textbf{3.1} && 2.8 \\
			
			HRT & 663.1 & 1.7 &&	186.3 & 1.7 && 157.8 & 1.6 && 138.0 & 1.5 &&  1.6 \\
			
			HRT (b$_{at}$=5, b$_{nat}$=5) & 811.0 & 1.4 &&	294.5 & 1.1 && 247.6 & 1.0 && 235.2 & 0.9 &&  1.1 \\
			
			HRT$_{12-1}$ & 249.6 & 4.5 && 111.5 & 2.8 && 85.1 & 2.9 	&& 83.9   & 2.4 && 3.1 \\
			HRT (k=3) & 448.7 & 2.5 && 134.8 & 2.3 && 111.7 & 2.2	&& 90.7  & 2.2 &&  2.3 \\
			HRT (k=4) & 360.0 & 3.1 && \textbf{111.4} & \textbf{2.8} && 85.8 & 2.9	&& 71.9  & 2.8 &&  2.9 \\

			\bottomrule[1pt]
		\end{tabular}
	}
	\vspace{-.5em}
	\caption{
	Compare the elapsed time and relative speedup ratio ($\alpha$) of decoding \entode \textit{newstest14} under different settings. 
	We use b$_{at}$=5, b$_{nat}$=1 and $k$=2 for HRT unless otherwise stated. HRT(b$_{at}$=5, b$_{nat}$=5) cannot decode data with batch size 32 (denoted by N/A) on GPU due to insufficient GPU memory. We bold the best results.
	}
	\label{table:speedup_wrt_model_varant}
	\vspace{-.5em}
	
\end{table*}

\section{AT Transformers in synthetic experiments}
\label{appendix:syn_exp}

\begin{table}[t]

	\centering
	\begin{tabular}{c  c c}
		\toprule[1pt]
		
		\textbf{AT Transformer} & \multicolumn{1}{c }{\textbf{En-Ro}} &
		\multicolumn{1}{c }{\textbf{En-De}} \\
		\midrule 
		
		\citet{vaswani2017attention}  & - & 27.3 \\
		\citet{ghazvininejad-etal-2019-mask} & 34.28 & 27.74 \\
		Our implementation & 34.25 & 27.45 \\
		\bottomrule[1pt]
	\end{tabular}
	\vspace{-.5em}
	\caption{The performance of autoregressive models in the synthetic experiment. }
	\label{table:at_baseline}
	\vspace{-.5em}
\end{table}
We trained all AT models in the synthetic experiment with the standard Transformer-Base configuration: layer=6, dim=512, ffn=2048, head=8. The difference from \citet{ghazvininejad-etal-2019-mask} is that they trained the AT models for 300k steps, but we updated 50k/100k steps on \entoro and \entode, respectively. Although fewer updates, as shown in Table~\ref{table:at_baseline}, our AT models have comparable performance with theirs. 

\section{Training algorithm}
\label{appendix:training_algo}

\begin{algorithm}[tb]
\caption{Training Algorithm for Hybrid-Regressive Translation}  
\label{alg:training}  
\begin{algorithmic}[1]
	\Require Training data $D$ including distillation targets, pretrained AT model $\textrm{M}_{at}$, chunk size $k$, mixed distillation rate $p_{raw}$, schedule coefficient $\lambda$
	\Ensure Hybrid-Regressive Translation model $\textrm{M}_{hrt}$
	
	\State $\textrm{M}_{hrt} \gets \textrm{M}_{at}$ \Comment{fine-tune on  pre-trained AT}
	\For {$t$ in $1,2,\ldots,T$} 
	\State   $\mX=\{\vx_1, \ldots, \vx_n\}$, $\mY=\{\vy_1, \ldots, \vy_n\}$, $\mY'=\{\vy'_1, \ldots, \vy'_n\}$ $\gets$ fetch a batch from $D$
	
	\For {$i$ in $1,2,\ldots,n$} 
	\State   $\mB_i=(\mX_i, \mY^*_i) \gets$ 
	uniformly sampling $\mY^*_i$ $\sim$ \{$\mY_i$, $\mY'_i$\} with $P(\mY_i)=p_{raw}$  \Comment{mixed distillation} 
	\EndFor
	
	\State   $p_{k} \gets (\frac{t}{T})^\lambda $ 
	\Comment{curriculum learning}
	
	\State   $\mB_{p}, \mB_{a} \gets \mB\big[:\lfloor n \times p_k \rfloor\big], \mB\big[\lfloor n \times p_k \rfloor:\big] $ \Comment{split batch}
	
	\State   $\mB_{p}^{at}, \mB_{p}^{nat} \gets$ construct training samples of primary tasks based on $\mB_{p}$ 
	\State   $\mB_{a}^{at}, \mB_{a}^{nat} \gets$ construct training samples of auxiliary tasks based on $\mB_{a}$
	
	\State   Optimize $\textrm{M}_{hrt}$ using $\mB_{p}^{at} \cup \mB_{a}^{at} \cup \mB_{p}^{nat} \cup \mB_{a}^{nat}$ \Comment{joint training} 
	\EndFor
\end{algorithmic}
\end{algorithm}

Algorithm~\ref{alg:training} describes the training process of HRT. The HRT model is pre-initialized by a pre-trained AT model (Line 1). 
Each training sample $\mB_i$ randomly selects a raw target sentence $\mY_i$ or its distilled version $\mY'$ (Line 4-6). 
Then according to the schedule strategy $p_k = \Big( \frac{t}{T} \Big)^\lambda$, we can divide $\mB$ into two parts: $\mB_{p}$ for primary tasks and $\mB_{a}$ for auxiliary tasks, where $|\mB_{p}|/|\mB|=p_k$ (Line 7-8). Next, we construct four kinds of training samples based on corresponding batches: $\mB_{p}^{at}$ (\texttt{TASK-SKIP-AT}), $\mB_{a}^{at}$ (\texttt{TASK-AT}), $\mB_{p}^{nat}$ (\texttt{TASK-SKIP-CMLM}) and $\mB_{a}^{nat}$ (\texttt{TASK-CMLM}). 
Finally, we collect all training samples together and accumulate their gradients to update the model parameters, which results in the batch size being twice that of standard training.

\definecolor{babyblue}{rgb}{0.54, 0.81, 0.94}
\definecolor{ballblue}{rgb}{0.13, 0.67, 0.8}
\definecolor{blue(munsell)}{rgb}{0.0, 0.5, 0.69}
\newcommand{\masktext}[1]{{\color{blue(munsell)}{\textbf{#1} \xspace}}}
\begin{table*}[t]
	{
		\centering
		\resizebox{\textwidth}{!}
		{
			\begin{tabular}{M{.15\linewidth} M{.75\linewidth}}
				\toprule[1pt]
				\textbf{Source} & 
				Also problematic : civil military jurisdiction will continue to be uph@@ eld . \\
				\midrule
				\textbf{Reference} &  
				
				Auch problematisch : Die zivile Militär@@ geri@@ chts@@ barkeit soll weiter aufrechterhalten bleiben . \\
				\midrule
				CMLM$_{10}$ (5th)
					& 
				\masktext{Problem@@} \masktext{atisch} \masktext{:} \masktext{Die} \masktext{zivile} \masktext{militärische} Gerichts@@ \uwave{barkeit} wird weiterhin \masktext{aufrechterhalten} \uwave{.} \eos  \\
				
				\midrule 
				
				HRT	& 
				\masktext{Auch} \uwave{problematisch} \masktext{:} \uwave{Die} \masktext{zivile} \uwave{Militär@@} \masktext{geri@@} \uwave{chts@@}  \masktext{barkeit} wird \masktext{weiterhin} \uwave{aufrechterhalten} \masktext{werden} \uwave{.} \masktext{\eos} \eos \\ 
				\bottomrule[1pt]
			\end{tabular}
		}
		\vspace{-.5em}
		\captionof{table}{A case study in \entode validation set. \masktext{Blue} denotes the original input is \mask. We add a wavy line under the target context tokens (black) that hit the reference translation.
		We also report the CMLM$_{10}$ in the 5th iteration as its masking rate is 50\%, closing to that of HRT.}
		\label{table:case_study}	
		\vspace{-.5em}
	}
\end{table*}

\section{Case study}
\label{appendix:case_study}
Table~\ref{table:case_study} shows a translation example from \entode validation set. Comparing CMLM and HRT, although both have the same masking rate (50\%), we can see two main differences: 
(1) The distribution of masked tokens in CMLM is more discontinuous than in HRT (see blue marks); 
(2) The decoder input of HRT contains more correct target tokens than CMLM, thanks to the Skip-AT decoder (see 
wavy marks). These two differences make our model easier to generate good translations than CMLM. It also indicates that our model is capable of generating appropriate discontinuous sequences.

\end{document}

%% file: math_commands.tex
\usepackage{amsmath,amsfonts,bm}

\def\eqref#1{equation~\ref{#1}}

\def\1{\bm{1}}

\def\vx{{\bm{x}}}
\def\vy{{\bm{y}}}
\def\vz{{\bm{z}}}

\def\mB{{\bm{B}}}

\def\mK{{\bm{K}}}

\def\mO{{\bm{O}}}

\def\mQ{{\bm{Q}}}
\def\mR{{\bm{R}}}

\def\mV{{\bm{V}}}

\def\mX{{\bm{X}}}
\def\mY{{\bm{Y}}}

\DeclareMathAlphabet{\mathsfit}{\encodingdefault}{\sfdefault}{m}{sl}
\SetMathAlphabet{\mathsfit}{bold}{\encodingdefault}{\sfdefault}{bx}{n}

